\documentclass[conference]{IEEEtran}
\IEEEoverridecommandlockouts
\usepackage{cite}
\usepackage{amsmath,amssymb,amsfonts}
\usepackage{algorithmic}
\usepackage{graphicx}
\usepackage{textcomp}
\usepackage{xcolor}
\usepackage{amsmath}
\usepackage{amssymb}
\usepackage{booktabs}
\usepackage[T1]{fontenc}
\usepackage{url}
\usepackage{stfloats}
\usepackage{amsfonts}
\usepackage{nicefrac}
\usepackage{microtype}
\usepackage{xcolor}
\usepackage{enumitem}
\usepackage{adjustbox}
\usepackage{latexsym}
\usepackage{tabularx}
\usepackage{caption}
\usepackage{subcaption}

\usepackage{url}
\def\BibTeX{{\rm B\kern-.05em{\sc i\kern-.025em b}\kern-.08em
    T\kern-.1667em\lower.7ex\hbox{E}\kern-.125emX}}
\begin{document}

\title{On Multi-Agent Deep Deterministic Policy Gradients and their Explainability for SMARTS Environment\\
}

\author{\IEEEauthorblockN{Ansh Mittal}
\IEEEauthorblockA{\textit{USC Viterbi School of Engineering)} \\
\textit{University of Southern California}\\
Los Angeles, CA, USA \\
anshm@usc.edu}
\and
\IEEEauthorblockN{Aditya Malte}
\IEEEauthorblockA{\textit{USC Viterbi School of Engineering)} \\
\textit{University of Southern California}\\
Los Angeles, CA, USA \\
malte@usc.edu}
}

\maketitle

\begin{abstract}
Multi-Agent RL or MARL is one of the complex problems in Autonomous Driving literature that hampers the release of fully-autonomous vehicles today. Several simulators have been in iteration after their inception to mitigate the problem of complex scenarios with multiple agents in Autonomous Driving. One such simulator--SMARTS, discusses the importance of cooperative multi-agent learning. For this problem, we discuss two approaches--MAPPO and MADDPG, which are based on-policy and off-policy RL approaches. We compare our results with the state-of-the-art results for this challenge and discuss the potential areas of improvement while discussing the explainability of these approaches in conjunction with waypoints in the SMARTS environment.
\end{abstract}


\section{Introduction}
\label{sec:1}
\begin{figure}[!ht]
\centering
\includegraphics[width=0.5\textwidth]{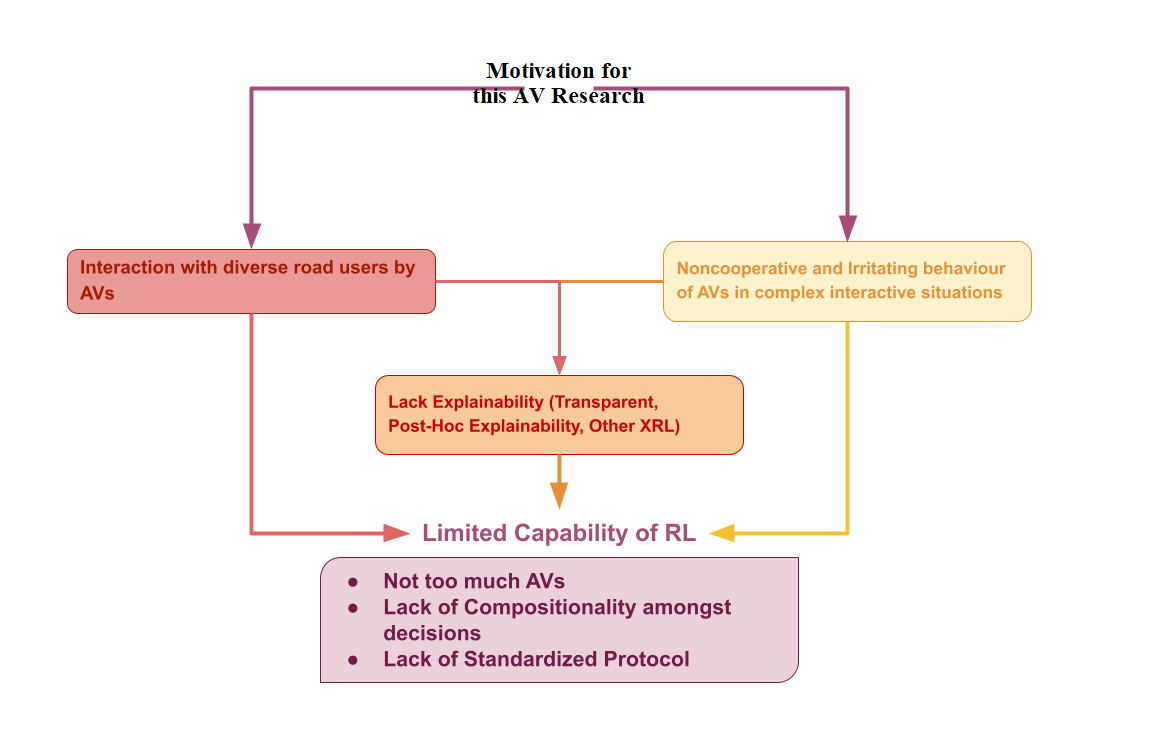}
\caption{Motivation}
\label{fig:1}
\end{figure}

We are interested in solving a multi-agent learning (MAL) problem related to the SMARTS environment~\cite{b1}. The main motivation for choosing this project was that despite the immense amount of deep learning architecture, the Autonomous Vehicle industry has yet to bring more Autonomous Vehicles on road. This majorly stems from the problems that arise when different agents interact with each other. This can further have a corresponding effect of the agents being non-cooperative with each other. Further, interpretability issues arise due to both of these which leads market stakeholders to invest less in the Autonomous Vehicle industry. This has been represented in fig.~\ref{fig:1}. Hence, till today there has been no fully automated vehicle on road. Further, only 54.2 million semi-Automated Vehicles~\footnote{Statistics taken from the website \url{https://www.statista.com/statistics/1230664/projected-number-autonomous-cars-worldwide/}} are projected to be there in 2024 which makes this a bleak future for the AV industry. Hence, we explore two different types of policies in the field of Multi-Agent Reinforcement Learning (MARL) and propose a way to work their explainability for future use-cases. We plan to open-source the code after extending and getting this research published. The simulations will be made available at that time as well.

\begin{figure*}[!ht]
\centering
\includegraphics[width=0.5\textwidth]{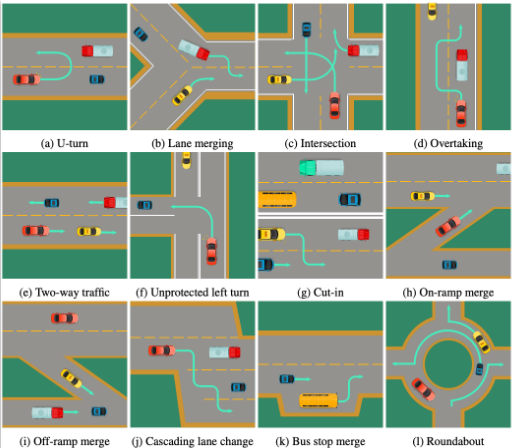}
\caption{Scenarios}
\label{fig:2}
\end{figure*}

The SMARTS (abb. Scalable Multi-Agent RL Training School) environment actually creates several scenarios such as the ones given in fig~\ref{fig:2}. It also defines different levels for different types of RL agents in Autonomous Vehicle literature as given below in table~\ref{tab:1}.

\begin{table*}
\centering
\scriptsize
\begin{tabular}{c @{\extracolsep{\fill}} cc}
\toprule 
\textbf{Level} & \textbf{Description} & \textbf{Possible RL Approaches}\\ 
\midrule
\textbf{M0} & Rule-based Planning & Not Applicable\\
\textbf{M1} & Single-Agent Learning w/o coordinated learning & Agent implicitly learns to anticipate other agent's actions according to our car and then takes the next steps\\
\textbf{M2} & Multi-Agent Modelling w Opponent Modelling & MARL to model learning of our agent using other agents, e.g., ``how likely to our car is to yield if their lane starts to change''\\
\textbf{M3} & Coordinated MAL w/ independent execution & Coordinated Learning of what to expect of each other even if no explicit coordination, e.g., some cars will leave some gap\\
\textbf{M4} & Local Equilibrium oriented MAL & Learn as a local group towards a certain equilibrium for double merge scenarios, etc.\\
\textbf{M5} & Social Welfare-oriented MAL & Learning broader repercussions of the car's actions\\
\bottomrule
\end{tabular}
\caption{Different levels of Multi-Agent Learning in Autonomous Driving discussed in the SMARTS environment~\cite{b1}}
\label{tab:1}
\end{table*}

The contributions of this project are three-fold:
\begin{enumerate}
\item We implement a Multi-Agent algorithm for Proximal Policy Optimization and Deep Deterministic Policy Gradients (using Prioritized Replay Buffers)
\item We compare their performance on the 4 evaluation metrics given in the SMARTS environment
\item Further, we discuss the explainability of the better algorithm for MARL
\end{enumerate}
The rest of the report is structured as follows. Section~\ref{sec:2} discusses the main problem statement of this research project effort. Section~\ref{sec:3} discusses the problems and challenges faced when working on this project. We discuss our solutions to this problem formulation in~\ref{sec:4}. We further list our results in section~\ref{sec:5} and compare them with the results obtained in the official challenge that NeurIPS organized. Then, we briefly discuss the explainability aspect that this problem can take from here on in section \ref{sec:6}. Conclusively, section \ref{sec:7} discusses the conclusions and reasons why our method falls short of the state-of-the-art results.

\section{Problem Definition}
\label{sec:2}
This project aimed to develop methods using different simulation environment data from SMARTS~\cite{b1} to innovate autonomous driving multi-agent that drive to their destination as safely and quickly as possible while following all the rules of the traffic system. For this formulation, the algorithms were evaluated on 4 different metrics, namely Completion, Time, Humanness, and Rules. All these metrics are discussed in detail in Section~\ref{sec:5}. This project leverages the SMARTS docker containers for simulations\footnote{\url{https://github.com/huawei-noah/SMARTS/tree/comp-1} (accessed: October 20, 2022)}. Further, a lot of different methods of environment setup (Docker, Singularity, and WSL direct setup) were tried for different reasons that have been mentioned in the next section.

\section{Technical Challenges}
\label{sec:3}
During working on this project, we faced several challenges which have been mentioned below.
\begin{enumerate}
    \item We were faced with the most striking requirement of GPU compute as the computation involved in any RL algorithm can take days to train on a normal CPU. For this, we had to further divide our approach into three steps.
    \begin{itemize}
    \item We first trained for 1000 episodes on the general system and ran a simulation while the model was training using Envision Sumo.
    \item Then, we took the model used for our approach to the Singularity on HPC.
    \item The model weights were then transferred to our system to check the model performance on the envision server since a GUI wasn't available on the HPC client.
    \end{itemize}
    \item Since there are multiple setups that can help us run the Envision Server, we were able to successfully run the competition environment but there were some files missing and version inconsistencies in different branches in the GitHub repository (till November 3, 2022; the competition end date ended up being postponed by a week due to this) and hence we weren't able to efficiently work at that time.
    \item With the baseline (Single Agent) that SMARTS proposes, there were issues with finding which approach works well. Further, the Ray package and TensorFlow packages can have conflicting dependencies for some versions. Hence, Version Control was vital for this research project.
\end{enumerate}

\begin{figure*}[!hbt]
\centering
\includegraphics[width=0.8\textwidth]{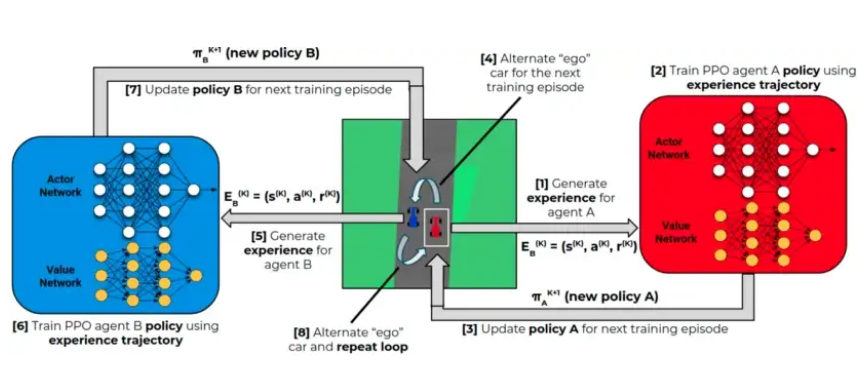}
\caption{Schematic Diagram for MAPPO using  Ray and TensorFlow environments (taken from the blog in the footnotes)}
\label{fig:3}
\end{figure*}
\begin{figure*}[!b]
\centering
\includegraphics[width=0.5\textwidth]{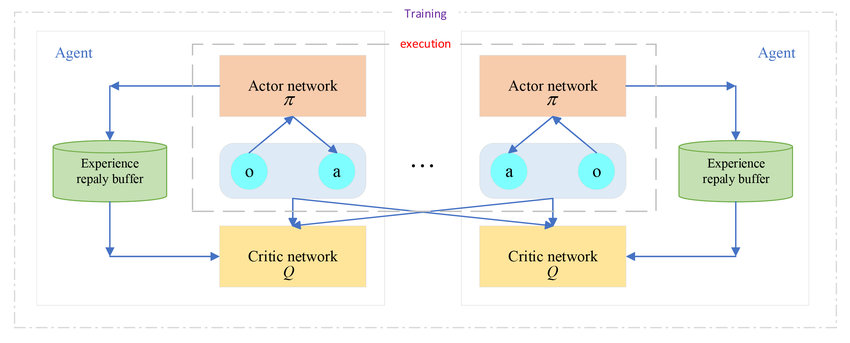}
\caption{Schematic Diagram for MADDPG taken from the research here~\cite{b14}}
\label{fig:4}
\end{figure*}

\section{Proposed Solution and Methodology}
\label{sec:4}
This section discusses the solution that the authors of this research project proposed for training the policies for the SMARTS environment. The authors used 2 different types of algorithms, namely, MAPPO (Multi-Agent Proximal Policy Optimization) and MADDPG (Multi-Agent Deep Deterministic Policy Gradient) for MAL and learning the policies. These have been discussed briefly below for the two algorithms mentioned in table~\ref{tab:2}.

\subsection{Multi-Agent Proximal Policy Optimization}
\label{sec:4.1}
Proximal Policy Optimization is a class of on-policy Reinforcement Learning algorithms that uses some of the advantages of Trust Region Policy Optimization (TRPO)~\cite{b2, b3, b4}. This set of algorithms alternates between sampling data from the environment through interactions, and optimizing a ``surrogate'' objective function that enables multiple epochs of minibatch updates. For more information about the algorithm, it is advised to read the original paper by OpenAI~\cite{b4}. Further, an adaptive clipping approach for PPO~\cite{b5} was developed later by building on prior works. After the inception of PPO, there have been various cooperative and Multi-Agent Proximal Policy Optimization implementations for various use cases such as targeted localization~\cite{b6}, Online scheduling for Production~\cite{b7}, Health-care~\cite{b8}. Moreover, there has been a combination of Convolutions with RL approaches such as CMAPPO (Convolutional Multi-Agent PPO)~\cite{b9} which learn an objective based on learning and exploring a new environment most effectively by combining various domains such as Convolutions (for RGBD+ information), Curriculum-based Learning, and Motivation-based Reinforcement Learning. Finally, there have been approaches that use the replay buffers approach from Deep Deterministic Policy Gradients (off-policy method)~\cite{b11} for the task of cooperative MARL\footnote{Github  source code can be found here \url{https://github.com/marlbenchmark/on-policy}}~\cite{b12}.
For implementing MAPPO, we referred to various sources to implement an architecture as given in the fig~\ref{fig:3}~\footnote{Blog read: \url{https://medium.com/analytics-vidhya/deep-multi-agent-reinforcement-learning-with-tensorflow-agents-1d4a91734d1f} and GitHub repository: \url{https://github.com/rmsander/marl_ppo.git}}. Further, OpenAI's Gym was used during the experiments for testing and tutorial purposes for understanding the various aspects of Multi-Agent Training and Self-Training for Open World Models~\cite{b13}\footnote{Github Page for the World Model which leverages MDN, CNN-VAE, and RL techniques for various RL-based simulations: https://worldmodels.github.io/}.

Apart from the approaches discussed above, there has been recent work on integrating the Actor-Critic method with the PPO RL paradigm that are beyond the scope of this project but should be able to guide this project in several future directions.

\subsection{Multi-Agent Deep Deterministic Policy Gradients (with Priority-based Replay buffers)}
\label{sec:4.2}
MADDPG is composed of various agents interacting cooperatively for MARL scenarios. It is composed of DDPG~\cite{b11} for multiple agents which interact with each other in a cooperative environment. It is an off-policy RL approach as opposed to MAPPO as it learns the value of the optimal policy independent of the agent's actions\footnote{Difference given here: \url{https://stats.stackexchange.com/questions/184657/what-is-the-difference-between-off-policy-and-on-policy-learning}}. Before discussing the term MAPPO, it should be interesting to get some background for Q-learning since DDPG is based on Q-learning~\footnote{Blog Post for introduction to Q-Learning: \url{https://medium.com/intro-to-artificial-intelligence/q-learning-a-value-based-reinforcement-learning-algorithm-272706d835cf}}. Since it's a model-free off-policy actor-critic algorithm that integrates the benefits of Deterministic Policy Gradients~\cite{b15} with the Deep Q-Networks (DQNs). Being an off-policy algorithm, it separately learns stochastic behavior policy for exploration and deterministic policy for target updates~\cite{b16}. Further, according to the original article, it maintains a replay buffer based on the temporal difference. But in our use case, we keep a priority list for this temporal replay buffer based on the number of accidents, rule violations, completion time, and jerks much like what had been done for MAPPO~\cite{b17}.\\
There has been a considerable extension to the MADDPG paradigm where recurrent DPGs were used for complex environments like Cognitive Electronic Warfare~\cite{b18} and Partially Observable Environments for Communication systems~\cite{b19}. A mixed environment approach was taken for complex environments using MADDPG~\cite{b20}, whereas Decomposed Approach was introduced for learning multi-agent policies for UAV clusters to build a connected communication network~\cite{b21}. Further, these set of algorithms were also discussed in the context of working with smart grids for edge technology~\cite{b22} and have shown to perform considerably well when compared to the other state-of-the-art.

\section{Results and Comparative Analysis}
\label{sec:5}
For the evaluation of the algorithms introduced in section~\ref{sec:4}, we discuss the evaluation metrics we calculated using the evaluation subdirectory on the branch comp-1 under track-1 of the NeurIPS competition. Please note that these approaches are discussed in the context of only online learning (which is listed in task 1 of the NeurIPS competition). So, the evaluation metrics are as given below.
\begin{enumerate}
    \item \textbf{Completion} This represents the completion of the total scenarios and hence can be calculated as the total number of crashes for each episode
    \item \textbf{Time} This metric discusses the total number of steps used by each agent (red cars) for each episode
    \item \textbf{Humannness} It refers to how close are our agents to imitating human-level driving scenarios. It is the average of the total distance to obstacles, angular jerk, linear jerk, and lane center offset for each episode
    \item \textbf{Rules} It refers to the total number of rules violated such as lane changing, Wrong Way, Speed Overlimit for each episode.
\end{enumerate}
It is of importance to note that since we are taking the average for each episode, there can be multiple agents (red cars) in a single episode. This leads to higher values in our evaluation and similar is the case with all other participating teams in the competition. We list our results with both the algorithms in table~\ref{tab:2}. 
\begin{table*}
\centering
\scriptsize
\begin{tabular}{ccccc}
\toprule 
\textbf{Algorithms} & \textbf{Completion} & \textbf{Time} & \textbf{Humanness} & \textbf{Rules}\\ 
\midrule
Speed-Direction Meta Controller Policies (TJUDRL lab) & \textbf{0.24 (1)} & \textbf{385.43 (1)} & 8161.24 (16) & 1.03 (12)\\
Waypoints-Trajectory Planning (CNN for Birdview) w/ Safety Score Learning & 0.31 (2) & 467.30 (4) & 4292.02 (12) & 16.96 (17)\\
Interaction-Aware Transformers & 0.32 (3) & 386.43 (2) & \textbf{641.86 (4)} & 0.43 (10)\\
\hline
\textbf{MAPPO (w Replay Buffer and fine-tuning)} & 0.72 (16*) & 723.1 (13*) & 4865.07 (13*) & 0.72 (12*)\\
\textbf{MADDPG (with Priority-based Replay Buffers \& fine-tuning)} & 0.64 (10*) & 746.29 (15*) & 1362.7 (9*) & \textbf{0.22 (9*)}\\
\bottomrule
\end{tabular}
\caption{Algorithms used and their comparison with the state-of-the-art winner solutions. The numbers in brackets are the ranks of the algorithm. The (*) represents projected ranks as we weren't able to submit our models to the official competition.}
\label{tab:2}
\end{table*}
It can be seen that MADDPG considerably outperforms MAPPO. This is because the priority experience buffers we explore for this problem statement take the gradients of speed, acceleration, and completion rate to define the priority queues. We further add a few screenshots for simulations for MADDPG in fig~\ref{fig:5}.

\section{On explainability of MADDPG}
\label{sec:6}
\begin{figure*}
\centering
\begin{subfigure}{.5\textwidth}
  \centering
  \includegraphics[width=0.7\linewidth]{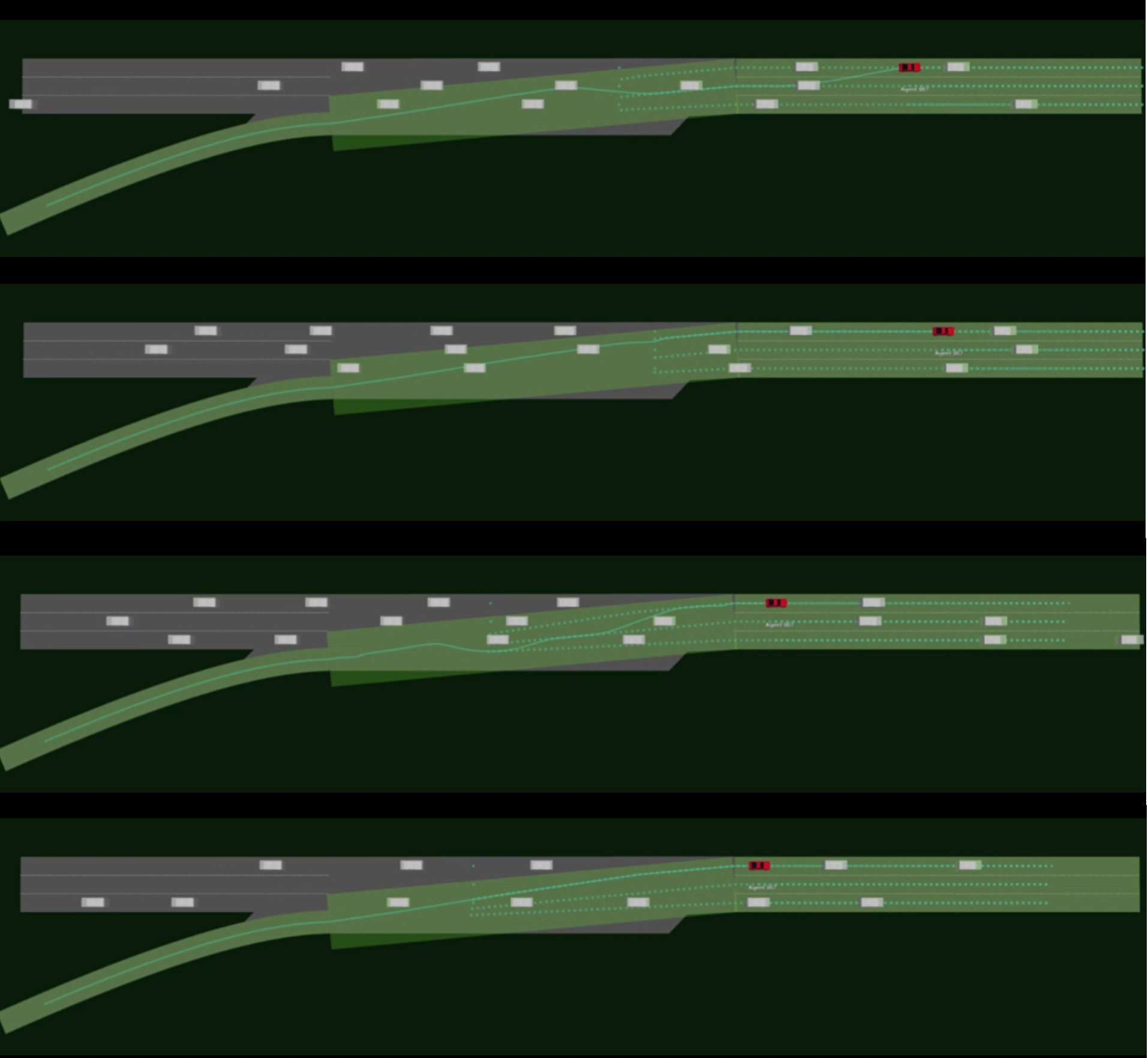}
  \caption{Highway Merge scenario}
  \label{fig:sub1}
\end{subfigure}%
\begin{subfigure}{.5\textwidth}
  \centering
  \includegraphics[width=0.4\linewidth]{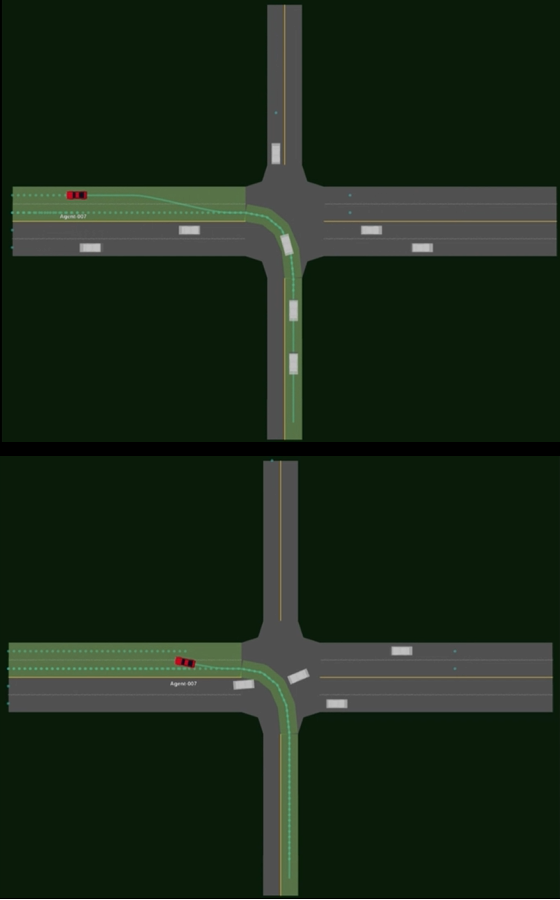}
  \caption{Zoo Intersection Scenario}
  \label{fig:sub2}
\end{subfigure}
\caption{Simulation results for MADDPG (w Priority Experience Replay Buffers)}
\label{fig:5}
\end{figure*}

We go through several research~\cite{b23, b24, b25} to comment on  the explainability of these deep reinforcement learning approaches to understand the decisions made by MARL policy networks. In our approach, we have used various metric parameters to design the priority experience replay buffer which can help the MADDPG learn the prioritized experience for better results. This is visible in the table~\ref{tab:2}, where we see that our MADDPG approach is able to outperform the top 3 winners of the SMARTS challenge. Further, we also observe better values for \textbf{Humanness} when compared to Top-2 winners. This happens as the priority experience replay buffer encodes the information for implicit explainability (transparent explainability) of the algorithm and post-hoc explainability of the model can be depicted by the blue waypoints as depicted in fig~\ref{fig:5}. This can help appeal to various stakeholders about how the car is taking action in the evaluation scenarios over different periods.

\section{Conclusion and Discussion}
\label{sec:7}
As noted above the results depict that our model falls short of the state-of-the-art approaches. This might be due to several reasons as discussed below.
\begin{enumerate}
\item The training for our model didn't converge when we stopped the training. We had to stop the training as the amount of time that we could query an HPC node was about to be exhausted and we had to save the results and get the evaluation scores for the same.
\item The top-2 approaches efficiently leverage Birdeyes view images from the simulation as feedback to change the directions of the car, which we haven't integrated into our model. This can lead to suboptimal performances.
\item Finally, suboptimal hyperparameters of the approaches that we leverage might lead to local minima rather than getting global minima.
\end{enumerate}
We plan to work on this project further and extend its scope twofold. First, we plan to implement a CV algorithm to reinforce the RL strategies with additional heuristics (left, right, keep) as is done in the best approach. Secondly, after getting feedback from the professor, we'd like to work on evolutionary computing for this approach. For example, these approaches can include Parameter-Exploring Policy Gradients with or without CV reinforcement~\cite{b26, b27}. We can also tune the hyperparameter search using evolutionary techniques like Genetic Algorithms, Particle Swarm Optimization, CMAES, Gray Wolf Optimization, etc (given enough GPU compute).

\section*{Acknowledgment}

We thank Prof Jyotirmoy V. Deshmukh for his continuous support during the duration of the course CSCI513. We also thank the course TAs--Xin Qin and Aniruddh Puranic, for their continuous support and accommodation during the duration of the course CSCI513. We thank USC for providing the services for High-Performance Computing for the purpose of this project.

% \vspace{12pt}
% \color{red}
% IEEE conference templates contain guidance text for composing and formatting conference papers. Please ensure that all template text is removed from your conference paper prior to submission to the conference. Failure to remove the template text from your paper may result in your paper not being published.


\begin{thebibliography}{00}
\bibitem{b1} Zhou, Ming, et al. "Smarts: Scalable multi-agent reinforcement learning training school for autonomous driving." arXiv preprint arXiv:2010.09776 (2020).
\bibitem{b2} Schulman, John, et al. "Trust region policy optimization." International conference on machine learning. PMLR, 2015.
\bibitem{b3} Kurutach, Thanard, et al. "Model-ensemble trust-region policy optimization." arXiv preprint arXiv:1802.10592 (2018).
\bibitem{b4} Schulman, John, et al. "Proximal policy optimization algorithms." arXiv preprint arXiv:1707.06347 (2017).
\bibitem{b5} Chen, Gang, Yiming Peng, and Mengjie Zhang. "An adaptive clipping approach for proximal policy optimization." arXiv preprint arXiv:1804.06461 (2018).
\bibitem{b6} Alagha, Ahmed, et al. "Target localization using multi-agent deep reinforcement learning with proximal policy optimization." Future Generation Computer Systems 136 (2022): 342-357.
\bibitem{b7} Lohse, Oliver, Noah Pütz, and Korbinian Hörmann. "Implementing an Online Scheduling Approach for Production with Multi Agent Proximal Policy Optimization (MAPPO)." IFIP International Conference on Advances in Production Management Systems. Springer, Cham, 2021.
\bibitem{b8} Allen, Ross E., et al. "Health-Informed Policy Gradients for Multi-Agent Reinforcement Learning." arXiv preprint arXiv:1908.01022 (2019).
\bibitem{b9} Sander, Ryan. "EMERGENT AUTONOMOUS RACING VIA MULTI-AGENT PROXIMAL POLICY OPTIMIZATION."
\bibitem{b10} Chen, Zichen, Budhitama Subagdja, and Ah-Hwee Tan. "End-to-end deep reinforcement learning for multi-agent collaborative exploration." 2019 IEEE International Conference on Agents (ICA). IEEE, 2019.
\bibitem{b11} Lillicrap, Timothy P., et al. "Continuous control with deep reinforcement learning." arXiv preprint arXiv:1509.02971 (2015).
\bibitem{b12} Yu, Chao, et al. "The surprising effectiveness of ppo in cooperative, multi-agent games." arXiv preprint arXiv:2103.01955 (2021).
\bibitem{b13} Ha, David, and Jürgen Schmidhuber. "World models." arXiv preprint arXiv:1803.10122 (2018).
\bibitem{b14} Qie, Han, et al. "Joint optimization of multi-UAV target assignment and path planning based on multi-agent reinforcement learning." IEEE access 7 (2019): 146264-146272.
\bibitem{b15} Silver, David, et al. "Deterministic policy gradient algorithms." International conference on machine learning. PMLR, 2014.
\bibitem{b16} Lillicrap, Timothy P., et al. "Continuous control with deep reinforcement learning." arXiv preprint arXiv:1509.02971 (2015).
\bibitem{b17} Hou, Yuenan, and Yi Zhang. "Improving DDPG via prioritized experience replay." no. May (2019).
\bibitem{b18} Wei, Xiaolong, et al. "Recurrent MADDPG for object detection and assignment in combat tasks." IEEE Access 8 (2020): 163334-163343.
\bibitem{b19} Wang, Rose E., Michael Everett, and Jonathan P. How. "R-MADDPG for partially observable environments and limited communication." arXiv preprint arXiv:2002.06684 (2020).
\bibitem{b20} Wan, Kaifang, et al. "ME‐MADDPG: An efficient learning‐based motion planning method for multiple agents in complex environments." International Journal of Intelligent Systems 37.3 (2022): 2393-2427.
\bibitem{b21} Zhu, Zixiong, et al. "Building a Connected Communication Network for UAV Clusters Using DE-MADDPG." Symmetry 13.8 (2021): 1537.
\bibitem{b22} Lei, Wenxin, et al. "MADDPG-based security situational awareness for smart grid with intelligent edge." Applied Sciences 11.7 (2021): 3101.
\bibitem{b23} Krajna, Agneza, et al. "Explainability in reinforcement learning: perspective and position." arXiv preprint arXiv:2203.11547 (2022).
\bibitem{b24} Wells, Lindsay, and Tomasz Bednarz. "Explainable ai and reinforcement learning—a systematic review of current approaches and trends." Frontiers in artificial intelligence 4 (2021): 550030.
\bibitem{b25} Heuillet, Alexandre, Fabien Couthouis, and Natalia Díaz-Rodríguez. "Explainability in deep reinforcement learning." Knowledge-Based Systems 214 (2021): 106685.
\bibitem{b26} Sehnke, Frank, et al. "Parameter-exploring policy gradients." Neural Networks 23.4 (2010): 551-559.
\bibitem{b27} Sehnke, Frank, et al. "Multimodal parameter-exploring policy gradients." 2010 Ninth International Conference on Machine Learning and Applications. IEEE, 2010.
\end{thebibliography}
\end{document}